\makeatletter
\@namedef{ver@everyshi.sty}{}
\makeatother
\documentclass[10pt,twocolumn,letterpaper]{article}
\pdfoutput=1
\usepackage{iccv}
\usepackage{times}
\usepackage{epsfig}
\usepackage{graphicx}
\usepackage{amsmath}
\usepackage{amssymb}
\usepackage{booktabs}
\usepackage{multirow}
\usepackage{multicol}
\usepackage{enumitem}
\usepackage{stfloats}
\usepackage{ulem}
\usepackage{changes}
\usepackage{cite}
\usepackage{authblk}

\usepackage[breaklinks=true,bookmarks=false]{hyperref}
\hypersetup{hidelinks}
\iccvfinalcopy 

\def\iccvPaperID{****} 

\ificcvfinal\fi

\begin{document}

\title{Knowledge Distillation from Single to Multi Labels: an Empirical Study}

\author[1]{Youcai Zhang}
\author[1,2]{Yuzhuo Qin}
\author[2]{Hengwei Liu}
\author[1]{Yanhao Zhang}
\author[1]{Yaqian Li}
\author[2]{Xiaodong Gu}
\affil[1]{OPPO Research Institute}
\affil[2]{Fudan University}

\maketitle
\ificcvfinal\thispagestyle{empty}\fi

\begin{abstract}
Knowledge distillation~(KD) has been extensively studied in single-label image classification. 
However, its efficacy for multi-label classification remains relatively unexplored. 
In this study, we firstly investigate the effectiveness of classical KD techniques, including logit-based and feature-based methods, for multi-label classification. 
Our findings indicate that the logit-based method is not well-suited for multi-label classification, as the teacher fails to provide inter-category similarity information or regularization effect on student model's training. 
Moreover, we observe that feature-based methods struggle to convey compact information of multiple labels simultaneously. 
Given these limitations, we propose that a suitable dark knowledge should incorporate class-wise information and be highly correlated with the final classification results. 
To address these issues, we introduce a novel distillation method based on Class Activation Maps~(CAMs), which is both effective and straightforward to implement.
Across a wide range of settings, CAMs-based distillation consistently outperforms other methods.
Code is available at \href{https://github.com/yzqinjacob/Distill-MLC}{https://github.com/yzqinjacob/Distill-MLC}.
\end{abstract}
\section{Introduction}
Obtaining highly accurate and lightweight deep neural networks is crucial in practice due to the limitations of computational resources.
To address this issue, Knowledge Distillation~(KD) has emerged as an effective technique.
By transferring knowledge from a high-capacity \textit{teacher} network to a relatively lightweight \textit{student} network during training, KD boosts the performance of student models.

KD has been extensively explored in recent years for computer vision tasks, such as image classification~\cite{hinton2015distilling,romero2014fitnets,zagoruyko2016paying,cho2019efficacy}, object detection~\cite{wang2019distilling,li2017mimicking,dai2021general}, and semantic segmentation~\cite{liu2019structured,wang2020intra}. 
The fundamental concept underlying KD is to identify appropriate forms of ``dark knowledge'' that can be effectively transferred from the teacher model to the student model. 
KD can be broadly classified into two important branches: logit-based and feature-based methods.
In the pioneering logit-based work~\cite{hinton2015distilling}, teacher models transfer the ``soft target'' containing the knowledge of similarity among different categories to enhance student models. 
Many influential feature-based works have also been proposed, represented by intermediate representations~\cite{romero2014fitnets} and attention map~\cite{zagoruyko2016paying}.

However, the literature lacks a comprehensive exploration of the application of KD in multi-label image classification, a more general and practical classification task, given that images are intrinsically multi-labeled.
This paper aims to fill this research gap by investigating the empirical mechanism of knowledge distillation for multi-label image classification. 
This paper seeks to address the following questions: \textit{1)}~What distinguishes single-label and multi-label distillation? \textit{2)}~Are classical distillation methods effective in multi-label classification? \textit{3)}~What constitutes an appropriate dark knowledge for multi-label distillation?

Generally, the pipeline is consistent between single-label and multi-label classification. 
The core difference is whether categories are mutually exclusive. Single-label classification predicts the unique label via \textit{Softmax} cross-entropy training. In contrast, multi-label classification is typically converted into a multiple binary classification problem via \textit{Sigmoid} cross-entropy training. Models can predict non-exclusive labels at the same time based on multiple independent binary one-vs-rest classifiers.

This paper investigates the limitations of existing methods in the multi-label setting. 
Our findings suggest that the classical logit-based~(soft target) distillation method does not typically work well for multi-label classification. 
The teacher model trained by multiple independent binary classifiers fails to provide information on similarities among categories or impose regularization, which are considered as the main reasons for the effectiveness of soft-target in single-label classification~\cite{hinton2015distilling,huang2022knowledge,bagherinezhad2018label,zhao2022decoupled,yuan2020revisiting}. 
Regarding feature-based distillation methods, we find they also have limitations for multi-label classification. They either suffer from redundant information~\cite{romero2014fitnets} or struggle to disentangle the knowledge of distinct labels~\cite{zagoruyko2016paying}.

To address these limitations and leverage both logit-based and feature-based methods, we propose class activation maps~(CAMs)~\cite{zhou2015cam} as an ideal form of dark knowledge for multi-label classification, which is both effective and straightforward to implement. 
CAMs represent discriminative class-specific regions, thus focusing well to different labels simultaneously, unlike the perception limitations of existing methods that focus on part labels.
Furthermore, CAMs are highly correlated with final classification results, since they are directly derived from the classifier. 

We conduct comprehensive experiments to test our hypotheses and empirically validate their rationality. Our results confirm that distillation with class activation maps~(CAMs) outperforms existing methods by a significant margin in various settings. Additionally, distillation with CAMs is versatile and can be applied in both full-label and missing-label settings of multi-label learning, as well as in single-label classification. Furthermore, we found that CAMs exhibit promising performance not only for CNN-based teacher models but also for transformer-based ones.

In summary, this paper presents an empirical and systematic study on knowledge distillation from single-label to multi-label classification. Specifically,
\begin{itemize}[topsep=-10pt,partopsep=0pt,itemsep=0pt,parsep=0pt]
    \setlength{\parskip}{0pt}
    \item [$\bullet$]
    We investigate the limitations of classical KD methods, including both logit-based and feature-based approaches, in multi-label classification. 
    \item [$\bullet$]
    We propose that dark knowledge in multi-label classification should incorporate class-wise information and be directly linked with classification results. To achieve this, we introduce a novel distillation method based on CAMs.
    \item [$\bullet$]
    We conduct extensive experiments to empirically verify the  rationality of the proposed points. 
    The superior performance on multiple datasets with various settings demonstrates the efficacy of distillation with CAMs.
\end{itemize}

\section{Related work}
\noindent\textbf{Knowledge Distillation~(KD)} was firstly proposed by Hinton \textit{et al.} in ~\cite{hinton2015distilling}.
KD aims at guiding the training of a smaller student network by transferring the knowledge from a teacher with stronger capacity. 
\textit{What knowledge} to be transferred is the hot research topic in KD, which roughly divide KD
into logit-based and feature-based distillation. Logit-based~\cite{hinton2015distilling} distillation measures Kullback-Leibler divergence of the soft target between teacher and student. Feature-based~\cite{romero2014fitnets,zagoruyko2016paying} distillation regards intermediate features as transferred cues. 
Besides these traditional works, an simple approach proposed in~\cite{yang2021knowledge} obtained promising improvement, which utilized teacher’s pre-trained classifier to train student’s penultimate layer feature. 
Method in~\cite{zhang2020prime} perceived the prime samples in distillation and then emphasized their effect adaptively. 
Recent research has also focused on interpreting the knowledge distillation mechanism~\cite{phuong2019towards,yuan2020revisiting,stanton2021does,zhao2022decoupled} and applying KD to various tasks~\cite{zhou2018rocket,chen2017learning,wang2019distilling,li2020block,salehi2021multiresolution}
However, a distillation study for multi-label classification is still missing.

\noindent\textbf{Multi-label image classification} is an important
and practical task in computer vision that can recognize multiple labels for a given image. 
The research direction on multi-label classification can be roughly categorized as: loss functions ~\cite{wu2020distribution,ridnik2021asymmetric,lin2017focal}, training scheme ~\cite{chen2019two,wu2016ml,zhang2018three}, classification head ~\cite{liu2021query2label,ridnik2023ml}.
Besides, label correlation modeling~\cite{yu2014multi,chen2019learning,chen2019multi,ye2020attention} and the utilization of region features\cite{wang2017multi,gao2021learning,narayan2021discriminative} are proved to be effective for multi-label classification.
In light of the challenge of annotating all ground-truth labels for an image, multi-label learning in the presence of missing labels~(MLML) has also attracted much research attention~\cite{yu2014large,cole2021multi,zhang2021simple,huang2019improving}.

\noindent\textbf{Class Activation Mapping~(CAM)} is a technique to obtain discriminative regions for specific classes in an image and generate class activation maps~(CAMs).
The original CAM~\cite{zhou2015cam} operates a weighted sum on the feature maps extracted by the backbone network.
It is restricted to networks with global average pooling~(GAP) layer. 
Grad-CAM~\cite{Selvaraju2017GradCAM} utilizes local gradient to generate CAMs in any architecture without the need for re-training.
It has been proven in \cite{Selvaraju2017GradCAM} that Grad-CAM and original CAM produce equivalent results in network with GAP layer. 
As GAP layer is a common structure in classification model, original CAM is suitable for most scenarios. Furthermore, there are also explorations on gradient-free extensions~\cite{Wang2020ScoreCAM,ramaswamy2020ablation}.
The CAM technique derived from classification network has been widely used for weakly supervised visual tasks, such as weakly supervised object location~\cite{gao2021ts,bai2022weakly,baek2020psynet} and object segmentation~\cite{liu2022partial,ahn2019weakly,wei2018revisiting}. In both tasks, category labels are employed as supervision, and CAM-based localization serves as supplementary information.

\section{Application of KD in Multi-label Learning}
\begin{figure*}[t]
\centering
\includegraphics[width=\textwidth]{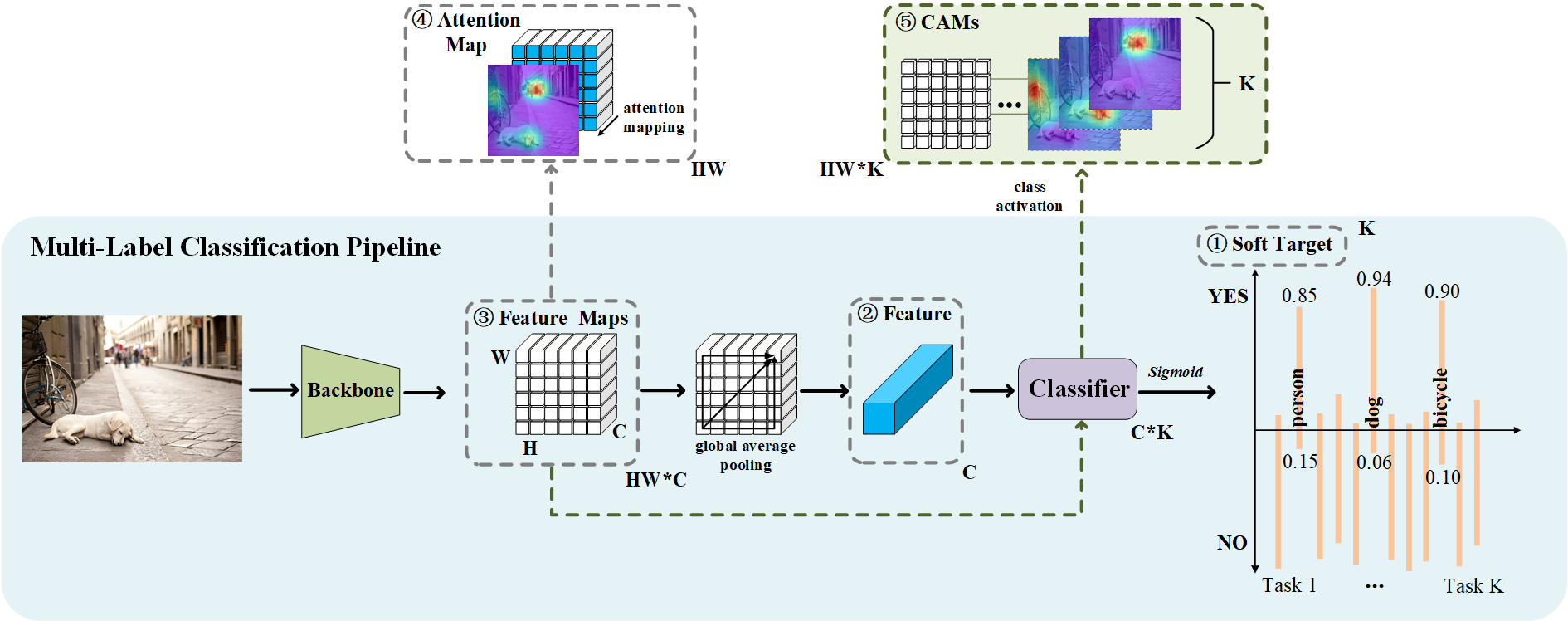}
\caption{Illustration of a typical multi-label classification pipeline and the acquisition of various knowledge for multi-label distillation. Grey dashed boxes represent the knowledge extracted by existing distillation methods, which can be categorized into Logit-Based(``\textit{\textcircled{\raisebox{-1.2pt}{1}} Soft Target}'') and Feature-Based(``\textit{\textcircled{\raisebox{-1.2pt}{2}} Feature}'', ``\textit{\textcircled{\raisebox{-1.2pt}{3}}Feature Maps}'' and ``\textit{\textcircled{\raisebox{-1.2pt}{4}} Attention Map}''). The green dashed one represents the proposed CAMs knowledge designed for multi-label distillation.
}
\label{pipline}
\end{figure*}

\subsection{Revisiting Existing Distillation Methods} \label{existing_method}
Distillation typically follows a paradigm that facilitates the student regress to the knowledge acquired from the teacher, meanwhile maintains the performance of the original task. Thus, the overall loss is a combination of task-specific~(\textit{e.g.}, classification) and distillation loss. It can be formulated as
\begin{equation}
    \mathcal{L} = \mathcal{L}_{cls} + \lambda \mathcal{L}_{distill},
\end{equation}
where $\lambda$ is a hyper-parameter to balance the classification and distillation losses. 

Before introducing KD, we first review the typical pipeline of multi-label image classification. 
As illustrated in Fig~\ref{pipline}, an input image is passed through the backbone network to generate \textit{Feature Maps}, which are then compressed into \textit{Feature} using global average pooling (GAP)~\cite{lin2013network} operation. Next, \textit{Feature} is fed into a linear classifier to obtain logits, and they are then transformed into multi-label classification probabilities using an activation function. 

Unlike single-label classification, multi-label classification recognizes multiple labels simultaneously.
To this end, a typical practice is to convert the multi-label problem into multiple independent binary problems that predict whether one label exists or not. 
The activation function in multi-label classification is \textit{Sigmoid} instead of \textit{Softmax}, because \textit{Softmax} yields a set of mutually exclusive probabilities for all categories while \textit{Sigmoid} produces independent binary predictions for each specific class.  

\subsubsection{Logit-Based Distillation}
Logit-based distillation adopts teacher's \textit{Soft Target} as the hint to guide the student. To obtain the \textit{Soft Target}, the output logits of classifier are softened using a temperature coefficient, followed by an activation function.
In single-label distillation, the Kullback–Leibler divergence is utilized to quantify the similarity of \textit{Soft Target} between teacher and student~\cite{hinton2015distilling}.
In multi-label distillation setting, we follow this approach by simply modifying the activation function for the \textit{Soft Target} generation from \textit{Softmax} to \textit{Sigmoid}.
Mathematically, we denote by $T_i^o$ and $S_i^o$ the classifier outputs of teacher and student respectively, then the logit-based distillation loss for multi-label distillation could be presented as
\begin{gather}
    T_i = \sigma(\frac{T_i^o}{\tau}), S_i = \sigma(\frac{S_i^o}{\tau}),\label{SoftTarget}\\
    \mathcal{L}_{KD} = {\tau}^2\sum_{i=1}^KKL(T_i||S_i)
\end{gather}
where $KL(\cdot)$ is the Kullback–Leibler divergence, $\tau$ is the temperature parameter, $K$ is the number of class and $\sigma(\cdot)$ represents the \textit{Sigmoid} function. $T_i$ and $S_i$ in Eq.~\ref{SoftTarget} are represented by ``\textit{\textcircled{\raisebox{-1.2pt}{1}} Soft Target}'' in Fig~\ref{pipline}.

\noindent\textbf{Limitation of \textit{Soft Target}:} \label{SoftTargetLimit}
\textit{Soft target in typical multi-label classification does not provide the similarity among categories or impose regularization on the student training.}

In the context of single-label classification, it is a general consensus that the success of soft-target distillation can be attributed to the information on similarities among different categories\cite{hinton2015distilling,huang2022knowledge,bagherinezhad2018label,zhao2022decoupled}.
Additionally, 
~\cite{yuan2020revisiting} has proved that \textit{Soft Target} is a type of learned label smoothing regularization by theoretical analysis, which also plays an important role in distillation.
However, in multi-label classification, each class is predicted independently, meaning that each \textit{Soft Target} contains only binary prediction for its own category and has no effect on other categories.
Thus the \textit{Soft Target} from teachers fails to provide additional information or impose regularization among categories.

Recently, a study\cite{zhao2022decoupled} has provided a novel explanation for soft-target  distillation by decoupling this method into target and non-target classes distillation. ``Target class" refers to the binary prediction of whether a certain class is present in the given image, while ``non-target" class refers to the similarity information among classes except for the target one. 
The study revealed that the non-target part makes the most contribution in soft-target distillation.
Nevertheless, \textit{Soft Target} in multi-label classification is characterized by a combination of multiple independent target classes, and lacks the critical non-target similarity information. 
Based on the above review, it is reasonable to conclude that the effectiveness of logit-based distillation may be limited in multi-label classification.

\subsubsection{Feature-Based Distillation}
Besides the \textit{Soft Target}, intermediate representations have also been proven to be effective knowledge, categorized as feature-based Distillation.
Considering the consistent processing flow with single-label classification, we can directly apply existing popular single-label distillation methods into multi-label learning.
Generally, the loss for feature-based distillation could be formulated as
\begin{gather}
    \mathcal{L}_{fea} = \sum d(T_{fea},S_{fea}),
\end{gather}
where $d(\cdot)$ indicates the distance function. $T_{fea}$ and $S_{fea}$ are the transferred knowledge from teacher and student respectively. The knowledge can be ``\textit{\textcircled{\raisebox{-1.2pt}{2}} Feature}''~\cite{zhang2020prime,yang2021knowledge}, ``\textit{\textcircled{\raisebox{-1.2pt}{3}} Feature Maps}''~\cite{romero2014fitnets} and ``\textit{\textcircled{\raisebox{-1.2pt}{4}} Attention Map}''~\cite{zagoruyko2016paying}, which are represented by grey dashed boxes in Fig~\ref{pipline}.
In our work, we discard \textit{Feature Maps} from intermediate layers of the backbone and only adopt the output of final block as the \textit{Feature Maps}.
The reason is that applying distillation on multiple layers performs worse than that on the end of the network, verified by \cite{yang2021knowledge}.
\textit{Attention Map} and \textit{Feature} is obtained by pooling operations on \textit{Feature Maps} along different dimensions. Mathematically, they can be represented as
\begin{align}
    &\textit{\textcircled{\raisebox{-1.1pt}{2}} Feature}: \sum\limits_{i \in HW}X_i,\\
    &\textit{\textcircled{\raisebox{-1.1pt}{3}} Feature~Maps}: X\in \mathbf{R}^{HW\times C},\\
    &\textit{\textcircled{\raisebox{-1.2pt}{4}} Attention~Map}: \sum\limits_{i \in C}X_i^2,
\end{align}
where $HW$ and $C$ denote spatial and channel-wise dimensions respectively.

Though \textit{Feature Maps} reserves relatively complete information for distillation, it is redundant and introduces extra storage usage.
\textit{Attention Map} can be considered as a refined knowledge of \textit{Feature Maps} by highlighting model's activation on a single map.
As another refined form of knowledge, \textit{Feature} adopts the output of the penultimate layer, it is directly connected to the classifier and has much impact on classification accuracy.
Thus, distillation with \textit{Feature} demonstrates superior performance among three feature-based methods in single-label classification~\cite{zhang2020prime,matsubara2021torchdistill,yang2021knowledge}.

\subsubsection{Comparison on ImageNet and COCO}
\label{ComparisonSingle2Multi}
Are existing classical distillation methods still effective in multi-label classification?
To answer this question, we apply above four methods into multi-label classification, and compare them on the representative benchmarks as shown in Table~\ref{tab:singleVSmulti}. 

We observe two intriguing results:
First, soft-target distillation obtain a non-negligible 0.84\% gain on COCO dataset, which is inconsistent with the explanation of its limitation discussed in Sec.~\ref{SoftTargetLimit}. 
This implies that the mechanism of \textit{Soft Target} in multi-label distillation needs further exploration. 
We empirically investigate and verify that the effectiveness of \textit{Soft Target} in multi-label distillation lies in the pseudo labels provided by the teacher, as discussed in Sec.~\ref{AnalysisOfSoftTarget}.
Second, feature distillation does not work well on COCO, unlike its superior performance on ImageNet. Interestingly, the inferior methods on ImageNet, such as \textit{Feature Maps} and \textit{Attention Map}, perform better on COCO. 
These differences suggest that the potential of existing feature-based methods has not been fully exploited.

\begin{table}[h]
  \centering
  \setlength{\tabcolsep}{3pt}
  \begin{tabular}{lcc}
    \toprule
    &top-1@ImageNet&mAP@COCO\\
    \midrule
    Teacher&73.30&82.40\\
    Student&69.76&74.09\\
    \midrule
    \textit{\textcircled{\raisebox{-1.2pt}{1}}} Soft Target~\cite{hinton2015distilling}&70.66&74.93 \\
    \textit{\textcircled{\raisebox{-1.2pt}{2}}} Feature~\cite{zhang2020prime,yang2021knowledge}&71.08&74.09\\
    \textit{\textcircled{\raisebox{-1.2pt}{3}}} Feature Maps~\cite{romero2014fitnets}&70.62&76.93\\
    \textit{\textcircled{\raisebox{-1.2pt}{4}}} Attention Map~\cite{zagoruyko2016paying}&70.69&76.28\\
    \bottomrule
  \end{tabular}
  \\ \hspace*{\fill} \\
  \caption{Comparison results of existing distillation methods on single-label~(ImageNet\cite{deng2009imagenet}) and multi-label~(COCO2014\cite{lin2014microsoft}) datasets. The student network is ResNet18 and the teacher networks are ResNet34 and ResNet101 respectively.}
  \label{tab:singleVSmulti}
\end{table}

\noindent\textbf{Limitation of feature-based methods:}
\textit{Existing Feature-based methods can not properly decouple the knowledge of different labels simultaneously.}

\begin{figure}[h]
\centering
\includegraphics[width=\linewidth]{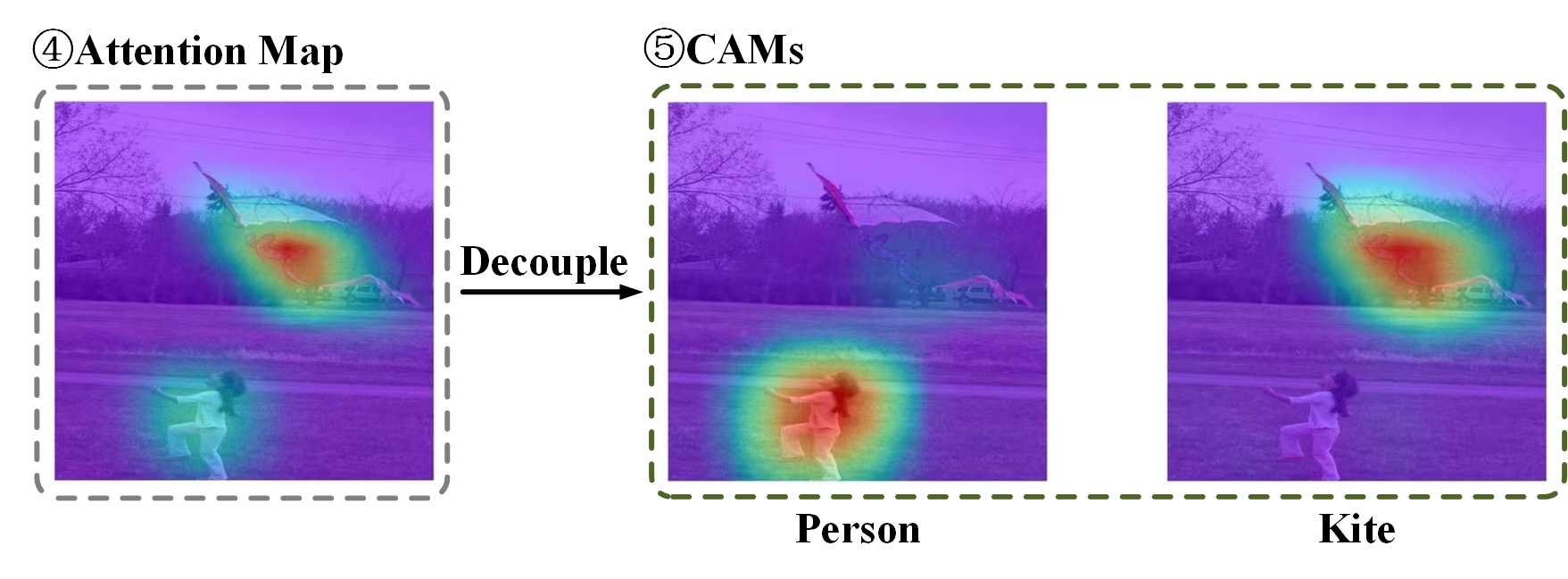}
\caption{Visualization of \textit{Attention Map} and \textit{CAMs}.
}
\label{decouple}
\end{figure}

We analyze above phenomenon of existing feature-based methods.
{\it{1)}} \textit{Feature} technique compresses \textit{Feature Maps} along the spatial dimension, resulting in the fusion of information from different instances, which leads to the incapacity of discrimination among different instances.
{\it{2)}} \textit{Attention Map} compresses \textit{Feature Maps} along channel-wise dimension and retains the spatial dimension, enabling distinguishable instances, as shown in Figure~\ref{decouple}. 
The retention of spatial dimension may explain its advantage over \textit{Feature} in multi-label distillation. 
However, \textit{Attention Map} still has limitations because it represents all instances on a single map, and its activation knowledge is agnostic to the corresponding category. 
And as shown in Figure~\ref{decouple}, \textit{Attention Map} can only focus well on a part of labels.
{\it{3)}} \textit{Feature Maps} 
 retains original information of different labels and performs better than \textit{Attention Map} on COCO.
But previous works~\cite{zagoruyko2016paying,liu2017learning} have proved that \textit{Feature Maps} is redundant and requires much storage usage.
Therefore, we argue that in multi-label distillation, the feature-based knowledge should convey compact information of multiple labels simultaneously.

\subsection{A Novel Distillation Method with \textbf{\textit{CAMs}}}
Through our revisiting of logit-based and feature-based methods, we have realized that the existing knowledge may not be the desired dark knowledge for multi-label classification.
Recently, a simple yet effective work~\cite{yang2021knowledge} also inspires us, which combines feature matching loss with the guidance of teacher’s classifier.
Therefore, we propose that the ideal dark knowledge should leverage both logit-based and feature-based methods.

For multi-label classification, we expect that \textit{the desired knowledge can perceive class-wise information and is directly linked with classification accuracy}.
Class Activation Maps~(\textit{CAMs}) naturally meet this requirement, as they reflect activation information from different categories and are also semantically informative, as these activation patterns are directly correlated with their corresponding logits.
For multi-label classification, we aim for the desired knowledge to have class-wise perception and a direct correlation with classification accuracy. Class Activation Maps~(\textit{CAMs}) naturally fulfill these desirable knowledge properties. 
\textit{CAMs} leverage the linear classification layer to generate attention maps for each class. 
Thus, we argue that \textit{CAMs} could be one of desired dark knowledges for multi-label classification.

In implementation, there are various methods for extracting \textit{CAMs}.
We adopt the original method proposed in~\cite{zhou2015cam}, as it can acquire the \textit{CAMs} simply and efficiently, requiring only one multiplication operation and no additional parameters.
Specifically, we denote the linear classifier by~$W \in \mathbf{R}^{C\times K}$. 
$W_i^k$ is the weight of the \textit{i-th} channel for classification logits of the \textit{k-th} category.
By performing a weighted-sum operation on \textit{Feature Maps}~$X\in \mathbf{R}^{HW\times C}$, we can obtain the \textit{k-th} class activating map $M_k \in \mathbf{R}^{HW}$.
$M_k \in \mathbf{R}^{HW}$ directly indicates the importance of the activation at spatial dimension correlated with class \textit{k}. This can be presented as
\begin{equation}\label{cam}
   \textit{\textcircled{\raisebox{-1.1pt}{5}}}~M_k = \sum_{i \in C}{W_i^k}X_i,
\end{equation}

To further exploit the classification information,
we adopt the teacher's predicted probabilities to re-weight the activation maps of different classes. 
By doing so, the distillation process can concentrate more on \textit{CAMs} that belong to the classes predicted to be positive for a particular class.
Mathematically, the loss function is denoted as
\begin{equation}\label{cam_loss}
   L_{CAM} = \sum_{k=1}^Kp_k d(M_k^T,M_k^S),
\end{equation}
where $p_k$ and $M_k$ represent the teacher's probability and activation map of the \textit{k-th} category. We use $L_2$ as distance function $d(\cdot)$ as it is simple and effective.

Distillation with \textit{CAMs} is effective in multi-label classification due to the following reasons:
{\it{1)}}~Visually, \textit{CAMs} decouples the \textit{Attention Map} into class-specific activation information, enabling it to attend well to different labels simultaneously, as shown in the second column of Figure~\ref{decouple}.
{\it{2)}}~\textit{CAMs} are highly correlated with final classification results and are rich in semantic information, as they are obtained through the classifier. The incorporation of teacher's predicted probabilities further enhances their effectiveness.
{\it{3)}}~The weighting term with teacher's probability plays a role of sample re-weighting in distillation. By assigning large weights to samples with high classification probability, the re-weighting process emphasizes the effect of easy samples, which has been validated to be appropriate in knowledge distillation in ~\cite{zhang2020prime}.

\section{Experiment} 
\subsection{Experimental Setting}
\noindent\textbf{Datasets.} 
We conduct experimentes on several classical datasets of different settings, including the general multi-label classification on MS-COCO\cite{lin2014microsoft}, PASCAL VOC 2007\cite{everingham2011pascal} and Open Images\cite{kuznetsova2020open}, missing-label setting of different ratios on MS-COCO, and singe-label classification on ImageNet\cite{deng2009imagenet}.
MS-COCO consists of 82,783 and 40,504 images for training and validation respectively, where 80 categories are covered. 
And following \cite{zhang2021simple}, we conduct distillation experiments on MS-COCO datasets with missing labels, which randomly drop positive labels for each training image with different ratios. 
VOC contains 5,011 training images with 20 categories and 4,952 images for test. 
Due to the challenge of downloading entire OpenImages dataset, we utilize its subset which consists of 1,742,125 training images and 37,306 test images, and 567 unique classes in total. 
ImageNet is a commonly used dataset which provides 1.2 million images from 1K classes for training and 50K for validation.

\noindent\textbf{Implementation.}
For multi-label learning, we use different backbones pre-trained on ImageNet~\cite{deng2009imagenet} for feature extraction. The input images are uniformly resized to $448 \times 448$. We train the model with Adam \cite{kingma2014adam} optimizer, True-weight-decay \cite{loshchilov2017decoupled} is set to $1e-2$, and cycle learning rate schedule \cite{smith2019super} is used with the max learning rate $1\times{10}^{-4}$. Besides, the exponential
moving average trick \cite{ridnik2021asymmetric} for better performance.
We adopt the asymmetric loss~\cite{ridnik2021asymmetric} as the classification loss. 
For single-label learning, the loss for classification is CrossEntropyLoss.

\noindent\textbf{Evaluation Metric.}
Following previous works, we adopt the mean average precision (mAP) over all categories for evaluation. For single-label learning, we report the Top-1 and Top-5 accuracies.

\subsection{Distillation from Single to Multi Labels}
\label{MissingLabelAnalysis}

\begin{table*}[t]
  \centering
  \begin{tabular}{l|ccccc}
    \toprule
    &&Full-label&75\%-label&40\%-label&single-label\\
    \midrule
    &Tea-ResNet101&82.40&78.76&72.43&70.82\\
    &Stu-ResNet18&74.09&70.03&63.60&60.14\\
    \midrule
    \multirow{2}{*}{Logit-Based}
    &Soft Target~\cite{hinton2015distilling}&74.93 &73.80 &69.83 &68.30 \\
    &Hard Target&74.56&73.34&69.98&69.27\\
    \midrule
    \multirow{2}{*}{Feature-Based}&Feature~\cite{zhang2020prime,yang2021knowledge}&74.09&72.07&66.98&65.20\\
    &Attention Map~\cite{zagoruyko2016paying}&76.28&72.53&67.22&65.58\\
    \midrule
    \multirow{2}{*}{CAMs-Based}&CAMs w/o tea-prob & 78.27 & 75.67 & 70.44 & 68.94 \\
    &CAMs&79.00&76.11&71.02&69.11\\
    \bottomrule
  \end{tabular}
  \\ \hspace*{\fill} \\
  \caption{mAP(\%) results of different approaches under different label missing ratios on COCO. For example, ``75\%-label" means that 75\% positive labels are retained in training set, and ``single-label" means that only one label is reserved for each image. Hard Target adds pseudo labels generated by teacher's soft target with a threshold.}
  \label{tab:missing}
\end{table*}

In Section \ref{existing_method}, we have observed that distillation methods exhibit different behaviors in single-label and multi-label classification. 
To investigate the distillation mechanism in multi-label classification further, we conduct experiments under a transitional form between single-label and multi-label learning, namely the \textit{multi-label learning with missing labels} setting. 
Following the approach in~\cite{zhang2021simple}, we retain only a certain ratio of positive labels, while the remaining labels are dropped as \textit{missing labels} during training on the COCO dataset. The validation set is fully labeled. Table \ref{tab:missing} presents the results of different distillation methods on various missing ratios of COCO.

\subsubsection{Analysis of Logit-Based Distillation}
\label{AnalysisOfSoftTarget}

Sec.~\ref{ComparisonSingle2Multi} presents a contradiction that needs to be resolved, \textit{i.e.}, why does soft target distillation still work in multi-label learning? 
Notably, Table \ref{tab:missing} shows that soft target distillation becomes more effective as label missing ratio increases. 
While under the full-label setting, soft label distillation only provides a 0.84\% gain, it provides up to an 8.16\% gain when the missing labels reduce it to a single-label form.
We hypothesize that the teacher network's stronger fitting ability makes it more robust to missing labels than the student network. 
Therefore, we believe that the teacher provides pseudo labels to the student through soft target in the case of missing labels, reducing the effect of false negatives from miss-labelling. 
To verify this, we transform the teacher's soft target into a binary hard label and use it as supervision to train the student network directly without distillation loss, which we refer to as "Hard Target" in Table~\ref{tab:missing}.

Table~\ref{tab:missing} shows that Hard Target achieves comparable or even better results than Soft Target, indicating that the most significant contribution of the teacher's soft target lies in the label correction of missing labels, rather than the similarity information among categories or regularization. We conclude that: 
\textit{The effectiveness of soft target in multi-label distillation lies in the pseudo labels provided by teacher.}

\begin{figure}[h]
\centering
\includegraphics[width=\linewidth]{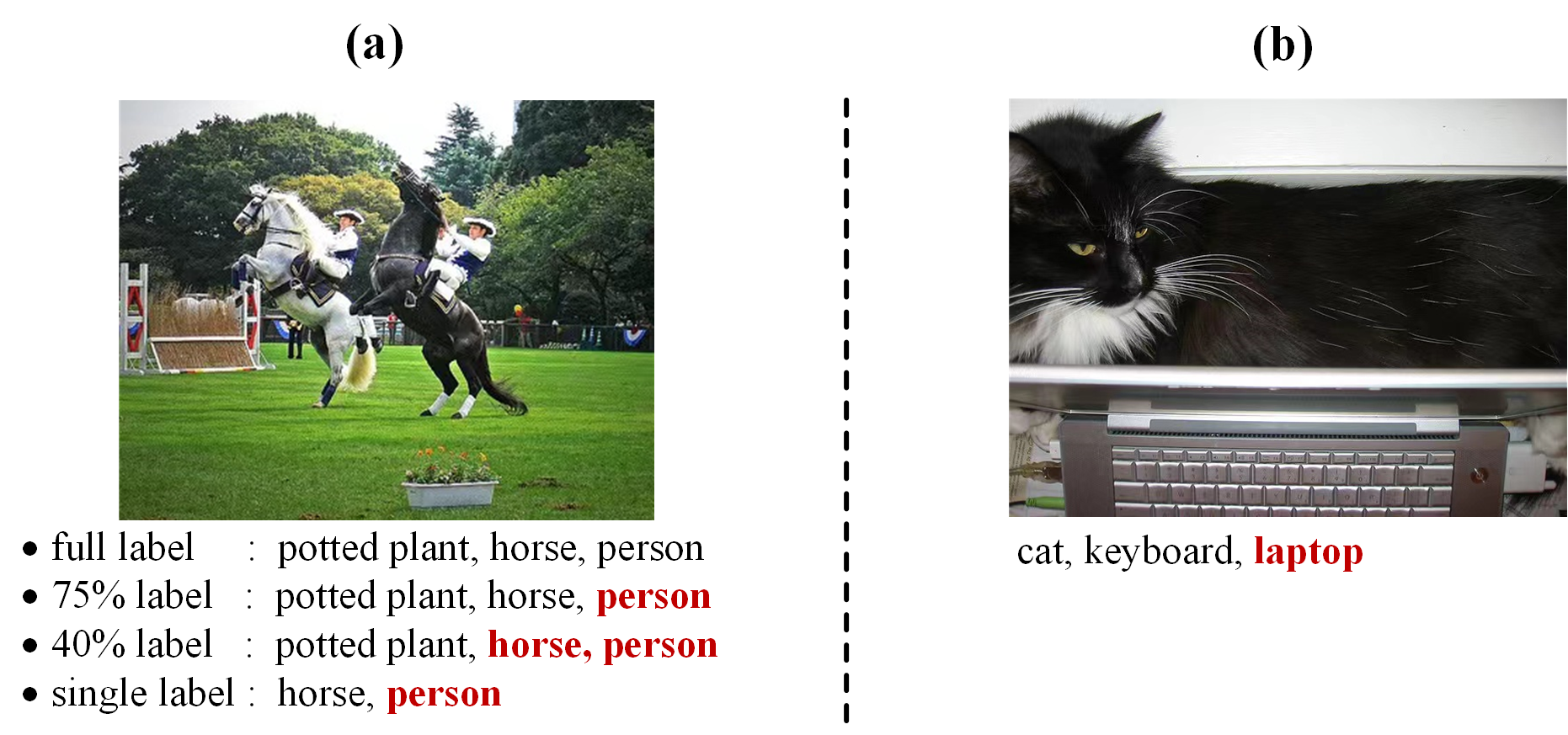}
\caption{Examples of pseudo labels on different missing ratios (a) and missing labels on full-label dataset (b). Pseudo labels and missing labels of two subfigures are both denoted with red color and bold font.
}
\label{pseudo}
\end{figure}

To better comprehend the label correction of missing labels, we present the pseudo labels added by teacher in the first line of Figure \ref{pseudo}. 
Besides, we observe that even on ``full-label" dataset, there are also ``missing-label" examples, as shown in the second line. 
This finding gives an explanation to the anomaly in Sec. \ref{ComparisonSingle2Multi}: soft target distillation benefits from teacher's pseudo label even on ``full-label" datasets that are not perfectly annotated. 

\subsubsection{Analysis of Feature-Based Distillation}
In addition to soft target, we also notice the impressive performance of feature under missing-label settings. 
To provide a comparison, we report results for attention map as well.
As shown in Table \ref{tab:missing}, the gap between feature and attention map narrows under missing-label scenarios compared to the full-label setting.

\begin{figure}[h]
\centering
\includegraphics[width=\linewidth]{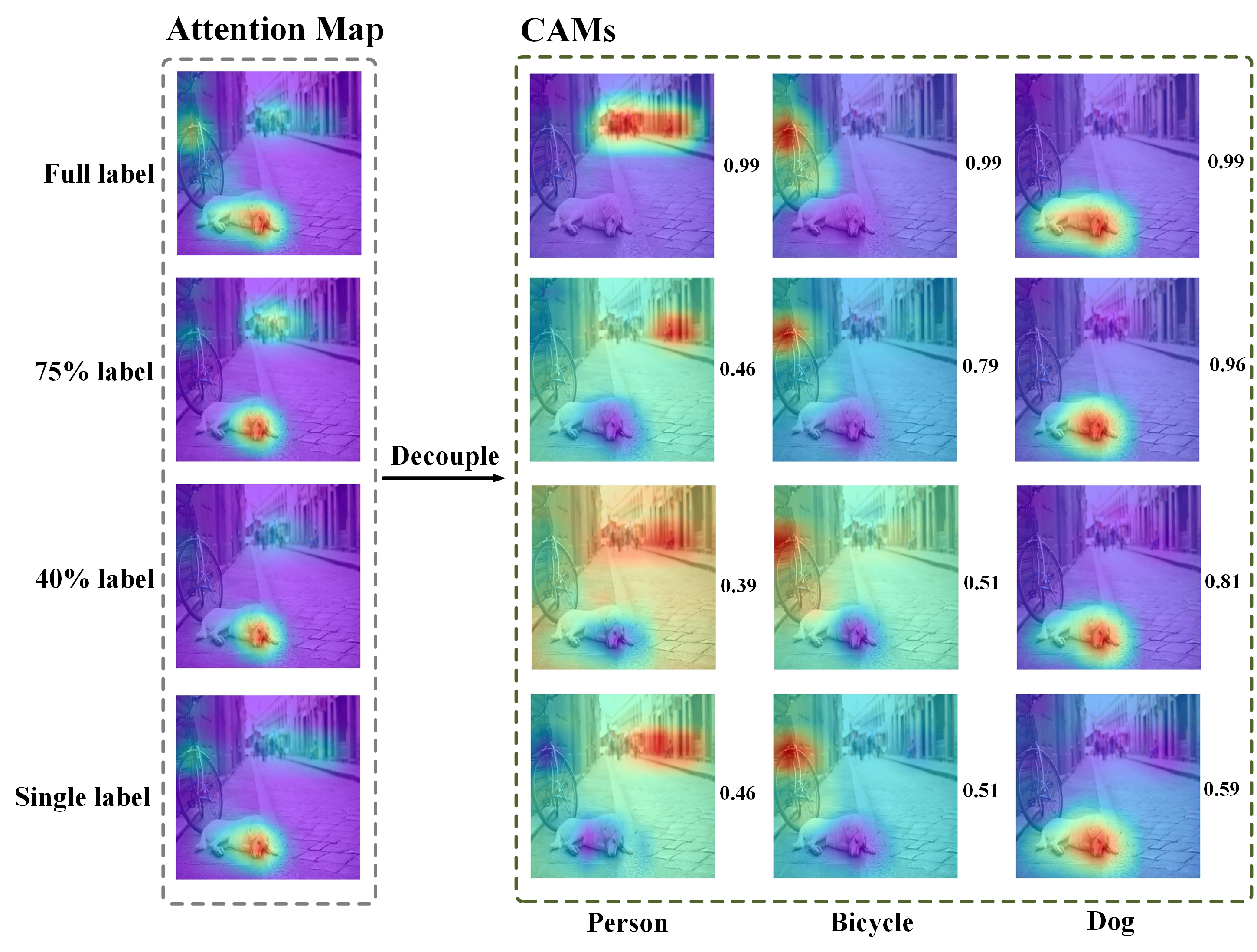}
\caption{Visualization of \textit{Attention Map} and \textit{CAMs} under different label missing ratios on COCO. Values on the right side of each picture represents the predicted probability. Attention map activates different instances in one map, while \textit{CAMs} can them.
}
\label{missing_cam}
\end{figure}

We conjecture that this is due to the model's inability to effectively decouple information, which may reduce the importance of the distillation knowledge's decoupling property.
To confirm this, we present Figure \ref{missing_cam}, which illustrates attention map under different missing ratios. 
It is evident that the model tends to focus on a single instance under missing-label settings, thereby weakening the importance of decoupling information. 
However, the gap between feature and attention Map is not eliminated, indicating that decoupled information is still valuable under missing-label settings. 
This motivates us to investigate how our \textit{CAMs} consisting of better decoupled knowledge works under missing-label scenes.

\subsubsection{Analysis of CAMs-Based Distillation}
Table~\ref{tab:missing} demonstrates that \textit{CAMs} distillation (without the teacher probability re-weighting strategy) consistently outperforms the feature-based distillation under both full-label and missing-label settings, and gradually performs better than the logit-based distillation as the label missing ratio decreases.
To qualitative analyze the benefits of \textit{CAMs} distillation, we present a few visualization examples in Figure~\ref{missing_cam}.
As demonstrated in Table~\ref{tab:missing}, the decoupling knowledge of different labels provided by \textit{CAMs} is crucial, regardless of whether it is under full-label or missing-label settings. 
Even though the model's decoupling ability drops under missing-label settings, \textit{CAMs} can still highlight the instances of corresponding labels, as shown in Figure~\ref{missing_cam}. 
The results demonstrate that using \textit{CAMs} to decouple attention maps is beneficial in multi-label distillation.

Table~\ref{tab:missing} also shows that \textit{CAMs} with the teacher's probability re-weighting strategy further improves the performance of \textit{CAMs} distillation. 
As shown in Figure~\ref{missing_cam}, the higher classification probability that the model assigns to a certain class, the more distinguishable instances in \textit{CAMs} are. 
The re-weighting process with the teacher's classification probability makes the student network focus more on prominent samples on which the teacher has much confidence. 
Thus, it further promotes the integration of classification information and the decoupling effect of \textit{CAMs}.

\begin{table*}[ht]
  \centering
  \begin{tabular}{cccccccc}
    \toprule
    & \multicolumn{2}{c}{COCO} & \multicolumn{2}{c}{VOC} & \multicolumn{2}{c}{OpenImages}\\
    \cmidrule(r){2-3} \cmidrule(r){4-5} \cmidrule(r){6-7}
    Teacher & ResNet101~\cite{2016Deep} & ResNet101 & ResNet101 & ResNet101 & ResNet101 & ResNet101\\
    Student & ResNet18~\cite{2016Deep} & MobileNetV3~\cite{DBLP:conf/iccv/HowardPALSCWCTC19} & ResNet18 & MobileNetV3 & ResNet18 & MobileNetV3\\
    \midrule
    Teacher & 82.40 & 82.40 & 92.56 & 92.56 & 68.89 & 68.89\\
    Student & 74.09 & 77.72 & 87.82 & 88.57 & 57.04 & 62.78\\
    \midrule
    Soft Target~\cite{hinton2015distilling} & 74.93 & 78.51 & 88.15 & 88.77 & 58.53 & 64.20\\
    Feature~\cite{zhang2020prime,yang2021knowledge} & 74.09 & 78.46 & 88.35 & 89.06 & 58.17 & 62.94\\
    Feature Maps~\cite{romero2014fitnets} & 76.93 & 78.25 & 88.49 & 88.77 & 57.71 & 63.51\\
    Attention Map~\cite{zagoruyko2016paying} & 76.28 & 78.66 & 90.21 & 89.62 & 59.05 & 63.78\\
    Feature+SR~\cite{yang2021knowledge} & 77.90 & 79.35 & 90.50 & 90.33 & 58.76 & 64.02\\
    \midrule
    \textbf{CAMs} & \textbf{79.00} & \textbf{80.41} & \textbf{90.98} & \textbf{90.38} & \textbf{59.35} & \textbf{64.97}\\
    \bottomrule
  \end{tabular}
  \\ \hspace*{\fill} \\
  \caption{mAP(\%) results of distillation with different backbones on three multi-label classification benchmarks~(COCO, VOC and OpenImages). Best results are represented as bold font.}
  \label{tab:mainresult}
\end{table*}

\subsection{Distillation on More Settings}
\subsubsection{Distillation on More Multi-Label Benchmarks}
To demonstrate the broad applicability of our proposed CAMs-based distillation method, we conduct distillation experiments on more multi-label benchmarks. 
In Table \ref{tab:mainresult}, we present the results obtained from various distillation methods on three multi-label datasets. 
The results confirm our analysis that classical single-label distillation methods are limited in multi-label distillation, as discussed in Section \ref{existing_method}, and our proposed \textit{CAMs} distillation method is more appropriate for multi-label distillation. Notably, our findings show that soft target distillation is particularly effective on OpenImages. This can be attributed to the fact that OpenImages is closer to a \textit{missing-label} dataset, given its vast scale compared to COCO and VOC datasets. As discussed in Sec. \ref{AnalysisOfSoftTarget}, in soft target distillation, students benefit more from pseudo-labels provided by the teacher as the label missing ratio increases.

\subsubsection{Distillation on the Single-Label Setting}
\setlength{\tabcolsep}{0.6pt}
\begin{table*}[t]
\begin{center}
\begin{tabular}{c|cc|ccccc|cccc}
\toprule
 & Tea-ResNet34 & Stu-ResNet18 & KD~\cite{hinton2015distilling} & L2~\cite{zhang2020prime,yang2021knowledge} & FitNet~\cite{romero2014fitnets} & AT~\cite{zagoruyko2016paying} & CAMs & CRD\cite{tian2019contrastive} & PAD-L2\cite{zhang2020prime} & KR\cite{chen2021distilling} & FM+SR\cite{yang2021knowledge}\\ 
\midrule
Top-1 & 73.30 & 69.76 & 70.66 & 71.08 & 70.62 & 70.69 & \textbf{71.55} & 71.17 & 71.71 & 71.61 & \textbf{71.73} \\
Top-5 & 91.42 & 89.08 & 89.88 & 90.19 & 90.01 & 90.13 & \textbf{90.34} & 90.51 & 90.45 & 90.51 & \textbf{90.60}\\
\bottomrule
\end{tabular}
\\ \hspace*{\fill} \\
\caption{Top-1 and Top-5 accuracy~(\%) on ImageNet. \textit{CAMs} distillation is compared with both classical methods and recent ones. The results of recent methods are reported from original papers.  \textit{Soft Target}, \textit{Feature}, \textit{Feature Maps}, and \textit{Attention Map} distillation in Table~\ref{tab:mainresult} are abbreviated as KD, L2, FitNet and AT respectively.}
\label{tab:imagenet}
\vspace{-1em}
\end{center}
\end{table*} 

Table~\ref{tab:missing} provides evidence that our proposed CAMs-based distillation method performs well under various missing label settings, which gives us confidence in its applicability to single-label learning classification tasks. To verify this, we applied \textit{CAMs} distillation to the influential ImageNet dataset. As shown in Table~\ref{tab:imagenet}, our proposed \textit{CAMs} distillation outperforms the four classical methods~\cite{hinton2015distilling,romero2014fitnets,zagoruyko2016paying} and is also competitive with recent single-label methods~\cite{yang2021knowledge,tian2019contrastive,chen2021distilling,zhang2020prime}. 
This demonstrates its versatility.
It is also worth noting that there are dataset-related factors that may impact the performance of different distillation methods on ImageNet. For example, as reported in \cite{yun2021re}, many images in the single-label ImageNet actually require multi-label annotations. This suggests that ImageNet could also be treated as a``missing-label" dataset, and the decoupling information provided by \textit{CAMs} is still effective in this context.

\subsubsection{Distillation with Transformer Teacher}
Transformers have shown remarkable success in various vision tasks. 
To exploit the advantages of both Transformer and CNN, we explore two distillation settings: Transformer-to-CNN and Transformer-to-Transformer distillation.
For Transformer-to-CNN distillation, we adopt SwinSmall~\cite{DBLP:conf/iccv/LiuL00W0LG21} as the teacher and ResNet18 as the student. For Transformer-to-Transformer distillation, we use SwinLarge~\cite{DBLP:conf/iccv/LiuL00W0LG21} and SwinTiny~\cite{DBLP:conf/iccv/LiuL00W0LG21} as the teacher and student respectively, in order to create a sufficiently large distillation gap. 
Table \ref{tab:swindistill} shows the results of our experiments, which demonstrate that distillation using CAMs also exhibits promising performance for Transformer-based teacher models.

Our experiments reveal that SwinSmall outperforms ResNet101 by 5.21\% in mAP on COCO, but results in a 1.85\% mAP drop when used as a teacher network for distillation with the ResNet18 student network. 
Previous works~\cite{cho2019efficacy,li2021reskd} have suggested that more accurate teachers are not always better for distillation, as small students may not be able to mimic large teachers due to capacity mismatches~\cite{cho2019efficacy}. 
We conjecture that the intrinsic structural differences between Transformers and CNNs could further exacerbate the difficulty of emulating the teacher model by the student model. 
Although our proposed CAM-based method does not completely solve the efficiency reduction problem of heterogeneous network distillation, it shows stronger adaptability compared to other methods.

\begin{table}[ht]
  \centering
  \begin{tabular}{ccc}
    \toprule
    Teacher & SwinSmall & SwinLarge \\
    Student & ResNet18 & SwinTiny \\
    \midrule
    Teacher & 87.61 & 90.43 \\
    Student & 74.09 & 85.08 \\
    \midrule
    Soft Target~\cite{hinton2015distilling} & 74.65 & 85.51 \\
    Feature~\cite{zhang2020prime,yang2021knowledge} & 74.35 & 85.93 \\
    Feature Maps~\cite{romero2014fitnets} & 75.60 & 85.50\\
    Attention Map~\cite{zagoruyko2016paying} & 74.45 & 85.24 \\
    Feature+SR~\cite{yang2021knowledge} & 74.50 & 86.09 \\
    \midrule
    \textbf{CAMs} & \textbf{77.15} & \textbf{87.45}\\
    \bottomrule
  \end{tabular}
  \\ \hspace*{\fill} \\
  \caption{mAP(\%) results of different methods using Transformer teachers on COCO dataset.}
  \label{tab:swindistill}
  \vspace{-1.5em}
\end{table}

\section{Conclusion}
This study presents a systematic and empirical study on knowledge distillation~(KD) from single-label to multi-label classification and offers a feasible solution that can improve the performance of small multi-label classifiers. 
Specifically, limitations of classical KD methods in multi-label classification are investigated and a novel distillation method based on Class Activation Maps~(\textit{CAMs}) is proposed.
The proposed method decouples the knowledge of different labels simultaneously and incorporates the final classification information, resulting in significant improvements over existing methods in various settings.

{\small
\normalem
\bibliographystyle{plain}
\bibliography{egpaper_final}
}

\end{document}